\documentclass[letterpaper, 10 pt, conference]{ieeeconf}  

\IEEEoverridecommandlockouts                              

\usepackage{graphics} 
\usepackage{epsfig} 
\usepackage{mathptmx} 
\usepackage{times} 
\usepackage{amsmath} 
\usepackage{amssymb}  
\usepackage{amsfonts}
\usepackage{hyperref}
\usepackage{verbatim}
\usepackage{subfigure}
\usepackage{multirow}
\usepackage{booktabs}
\usepackage{wrapfig}
\usepackage[english]{babel}
\usepackage{algorithm}
\usepackage{algorithmic}
\usepackage{verbatim}
\usepackage{stfloats}

\title{\LARGE \bf
Deep Reinforcement Learning for Localizability-Enhanced Navigation in Dynamic Human Environments
}

\author{Yuan Chen, Quecheng Qiu, Xiangyu Liu, Guangda Chen, Shunyi Yao, Jie Peng, Jianmin Ji$^*$ and Yanyong Zhang$^*$
\thanks{Authors are with University of Science and Technology of China, Hefei 230026, China.
Jianmin Ji and Yanyong Zhang are corresponding authors {\tt\small \{jianmin,yanyongz\}@ustc.edu.cn}.}
}

\begin{document}

\maketitle
\thispagestyle{empty}
\pagestyle{empty}

\begin{abstract}
Reliable localization is crucial for autonomous robots to navigate efficiently and safely. 
Some navigation methods can plan paths with high localizability (which describes the capability
of acquiring reliable localization).
By following these paths, the robot can access the sensor streams that facilitate more accurate location estimation results by the localization algorithms.
However, most of these methods require prior knowledge and struggle to adapt to unseen scenarios or dynamic changes. To overcome these limitations, we propose a novel approach for localizability-enhanced navigation via deep reinforcement learning in dynamic human environments.
Our proposed planner automatically extracts geometric features from 2D laser data that are helpful for localization. The planner learns to assign different importance to the geometric features and encourages the robot to navigate through areas that are helpful for laser localization. To facilitate the learning of the planner, we suggest two techniques: (1) an augmented state representation that considers the dynamic changes and the confidence of the localization results, which provides more information and allows the robot to make better decisions, (2) a reward metric that is capable to offer both sparse and dense feedback on behaviors that affect localization accuracy.
Our method exhibits significant improvements in lost rate and arrival rate when tested in previously unseen environments. 
\end{abstract}

\section{INTRODUCTION}
The capability of a robot to achieve reliable localization is a fundamental skill for autonomous navigation systems. In many applications, incorrect localization results can lead to bad effects on navigation performance and safety. For example, a delivery robot may fail to deliver its goods to the correct location, and a shopping guide robot may fall down an escalator when it runs out of its work region. Additionally, the effect can be much more severe for search and rescue robots, where human lives may be at risk. However, the robot will inevitably be subjected to localization errors or uncertainties due to various factors, such as sensor noise, environmental changes, and mechanical drift. These errors can accumulate over time in complex dynamic environments and lead to \emph{navigation lost}, where the robot loses track of its position and orientation in the environment~\cite{fox1999monte}.
To overcome these issues, some navigation methods plan paths not only to be collision-free but also to provide access to a sensor stream that can obtain more reliable localization along the path. 
However, most of these localizability-enhanced methods require prior knowledge, such as predefined patterns of landmarks~\cite{zhang2018perception,bartolomei2020perception} or pre-built maps that quantify the localizability of each region in the environment with a specific metric (i.e., the entropy or the pose estimation covariance)~\cite{roy1999coastal,schirmer2017efficient,gao2019new}. 
Based on this prior knowledge, these methods can achieve considerable performance in restricted environments. However, it is challenging to generalize these methods well to unforeseen scenarios with unfamiliar classes of landmarks or adapt these methods well to dynamic changes that cannot be recorded in pre-built maps.

This paper aims to develop a novel localizability-enhanced navigation approach in dynamic human environments without requiring extensive prior knowledge. 
To achieve this goal, the main challenges we need to overcome are twofold: (1) how to extract features conducive to localization from raw sensing data and assign them appropriate weights for planning and (2) how to interact with pedestrians and balance navigation efficiency, obstacle avoidance safety, and localization robustness.

To address the above challenges, we propose a novel approach that utilizes deep reinforcement learning (DRL). Our approach takes advantage of Deep Neural Network (DNN) to automatically extract geometric features from 2D laser data that are helpful for localization.
The DNN-based policy adapts to the scene online and assigns appropriate importance to the extracted geometric features, achieving a balance between efficiency, safety, and localization robustness through trial and error in Reinforcement Learning (RL).
To the best of our knowledge, this is the first navigation method that enhances localizability in dynamic human environments without relying on predefined landmark features or localizability maps.
To facilitate the agent to learn the desired policy, we meticulously designed the representation of the state space and reward function. We augment the state space by incorporating pedestrian maps~\cite{yao2021crowd} to track the movement of surrounding pedestrians and include pose estimation variance to measure the quality of localization results. This allows the robot to make more informed decisions that consider the dynamic changes in the environment and the confidence in the localization results. Moreover, we define a reward function that can provide sparse feedback on \emph{navigation lost} and dense feedback on behaviors affecting localization accuracy. It is important to note that the ground truth of the robot pose is only available during training, and the robot must rely on estimated poses during execution. We also design specialized pedestrian scenarios and policy network architectures for training and successfully train a localizability-enhanced navigation policy that can constantly obtain more reliable localization results.
 
For the purpose of evaluating the effectiveness of our proposed approach, we established a set of previously unseen environments in a simulation as a testing benchmark. A range of metrics was utilized to assess the performance of our method. The experimental results show that our approach achieves a loss rate that is one order of magnitude lower than that of other navigation methods while simultaneously increasing the arrival rate by $25\%$. This performance improvement comes at the cost of an $9\%$ increase in moving distance. Additionally, we provide a visualization of the activation of the first laser feature layer to demonstrate the efficacy of our approach in automatically extracting features. Finally, we conducted a real-world deployment of our trained policy on a physical robot to show its practicality.
Our main contributions are summarized as follows:
\begin{itemize}
	\item We propose a DRL-based navigation method that enhances localizability in dynamic human environments.
	\item To learn the behaviors that enhance localizability, it is essential to incorporate information about localization quality into the robot's state and use a reward function that gives frequent feedback on the behaviors that affect localization.
        \item We visualize the activation of the first laser feature layer to show the geometry features extracted to make localizability-enhanced decisions.
        \item We verify the generalization of our method through simulation testing in various scenarios and the deployment of a physical robot.
\end{itemize}

\section{Related Work}
The \emph{navigation lost} problem occurs when the robot's localization uncertainty or error becomes too large during navigation, causing the robot to lose track of its position and orientation in the environment~\cite{fox1999monte}. 
Past solutions to this problem have been primarily passive, assuming that the robot's motion and sensor direction cannot be controlled. These methods selectively use the sensor stream to minimize localization uncertainty or error~\cite{fox1999markov, tipaldi2013lifelong, sun2016towards}.
However, if the localization routine has partial or full control over the robot's movement and sensor direction, it can increase the efficiency and robustness of localization~\cite{burgard1997active,borghi1998minimum}, particularly in dynamic environments where moving people may occlude the sensors' view~\cite{li2016active,fan2019getting}. 
These active localization algorithms focus on the global localization problem where the initial position of the robot is unknown and usually works when \emph{navigation lost} occurs. 

With the intention of avoiding \emph{navigation lost}, previous work such as~\cite{roy1999coastal,schirmer2017efficient,gao2019new} consider the localizability in path planning.
They generate a localizability map that contains the information content of each position in the environment. Path planning is then optimized over both conventional costs, such as distance or time, and localization certainty using the localizability map. 
Based on prior knowledge, they can make long-range plans with a global view but have trouble adapting well to dynamic changes through real-time perception.

To make use of real-time perception, other prior knowledge types have been introduced. Prediction-based methods~\cite{sim2005global, costante2018exploiting} predict the future pose estimated error and uncertainty based on specific pose propagation rules and then use the predicted error or uncertainty as one of the optimization items for path planning. Landmark-based methods recognize particular classes of landmarks from sensor data with a pre-designed pattern~\cite{zhang2018perception} or DNN-based semantic segmentation~\cite{bartolomei2020perception}. They can evaluate candidate paths based on landmark concentration or the category of the landmark in view. However, these methods rely heavily on prior knowledge and are often difficult to generalize to unforeseen environments.

For the purpose of improving generalizability, DRL has been applied to localizability-enhanced navigation. 
Taking visual images as inputs, \cite{bartolomei2021semantic, bartolomei2022autonomous} present semantic-aware path-planning pipelines in which RL agents utilize semantic masks to generate actions.
To the best of our knowledge, there is currently only one DRL-based localizability-enhanced navigation algorithm that directly feeds raw sensing data (used for localization) to the RL agent, which is \cite{lin2022localisation}.
However, \cite{lin2022localisation} does not consider the interaction with pedestrians, which limits their usefulness in real-world scenarios. 
As the implementation details of \cite{lin2022localisation} are not presented, we did not compare them with us in our experiments.

\begin{figure*}[tp]
	\centerline{\includegraphics[width=0.84\linewidth]{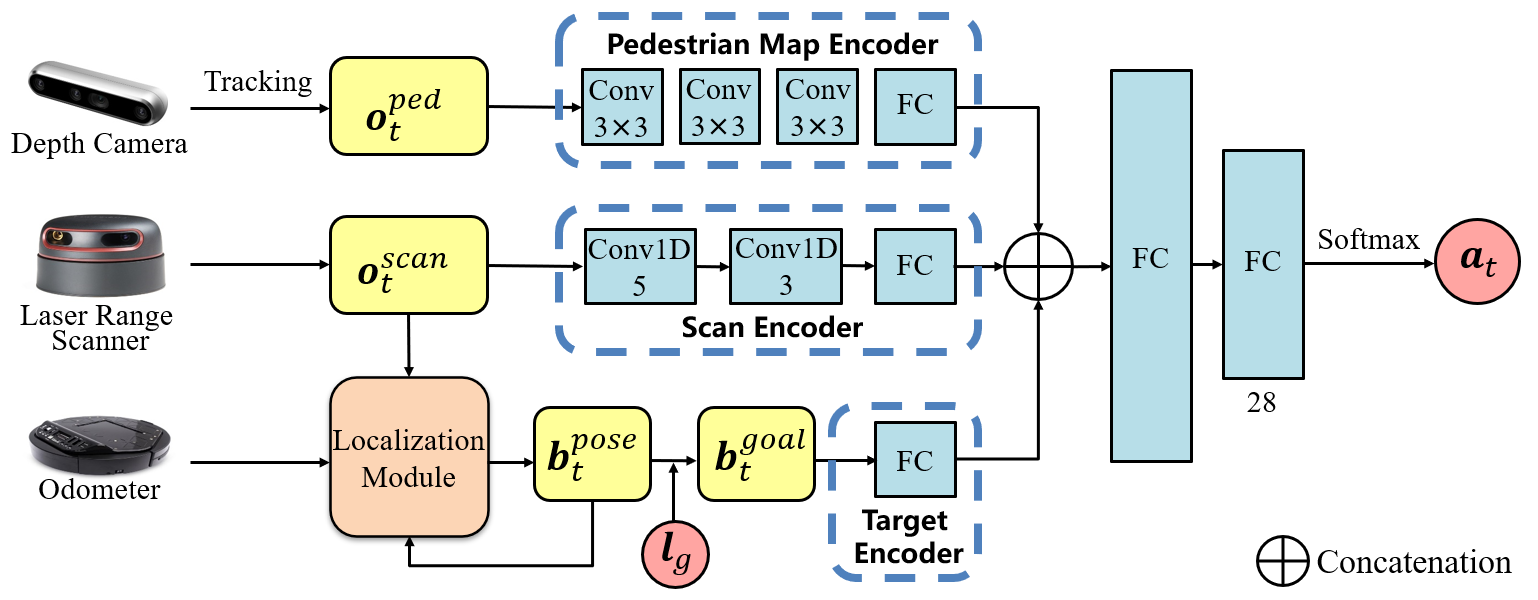}}
	\caption{The architecture of the policy network combined with a third-party localization module. The input of the network includes the pedestrian map $\mathbf{o}_t^{ped}$, the scan measurements $\mathbf{o}_t^{scan}$, and the belief over the relative goal position $\mathbf{b}_t^{goal}$, which is calculated by the location of the goal $\mathbf{l}_g$, and the belief over the robot current pose $\mathbf{b}_t^{pose}$.
	$\mathbf{b}_t^{pose}$ is estimated by the localization module with scan measurements, odometer data, global map, and $\mathbf{b}_{t-1}^{pose}$.
	The network uses a pedestrian map encoder, a scan encoder, and a target encoder to process $\mathbf{o}_t^{ped}$, $\mathbf{o}_t^{scan}$, and $\mathbf{b}_t^{goal}$, respectively. Then, the network concatenates the outputs of the encoders, feed them into two fully connected (FC) layers, and gets a pair of linear and angular velocities, i.e, $\mathbf{a}_t$.}
    \label{fig2}
\end{figure*}

\section{Problem Formulation}
The problem of localizability-enhanced navigation in dynamic human environments can be viewed as a sequential decision-making problem that aims to reach a target location~$\mathbf{l}_g$ while satisfying both collision-free and localization error constraints.
At each time step $t$, given a frame sensing data~$\mathbf{o}^{sensor}_t$, odometer data $\mathbf{o}^{odom}_t$, and the probability distribution of the robot pose estimated at the previous time step $\mathbf{b}^{pose}_{t-1}$, a third-party localization module $f$ estimates the probability distribution of the robot pose at the current time step $\mathbf{b}^{pose}_{t}$, i.e.,
$$\mathbf{b}^{pose}_{t}=f(\mathbf{o}^{sensor}_t, \mathbf{o}^{odom}_t, \mathbf{b}^{pose}_{t-1}).$$
The probability distribution of the
goal position in the robot coordinate system $\mathbf{b}^{goal}_t$ can be computed with $\mathbf{b}^{pose}_{t}$ and $\mathbf{l}_g$.
Then, a navigation policy $\pi_\theta$ provides an action command based on the motion data of the nearby pedestrians $\mathbf{o}^{ped}_t$, $\mathbf{o}^{sensor}_t$, and $\mathbf{b}^{goal}_{t}$ as follows:
$$ \mathbf{a}_t=\pi_\theta(\mathbf{o}^{ped}_t, \mathbf{o}^{sensor}_t, \mathbf{b}^{goal}_{t}),$$
where $\theta$ is the model parameters.
Our goal is to find a policy to minimize the expectation of the arrival time while accounting for the robot's safety by ensuring that it does not collide or get lost. We define getting lost as:
$$\Vert \mathbf{p}^{gt}_t - \mathop{arg\max}_\mathbf{p} \mathbf{b}^{pose}_t(\mathbf{p})  \Vert > \varepsilon,$$
where $\mathbf{p}^{gt}_t$ denotes the ground truth of the robot pose at $t$, $\mathop{arg\max}_\mathbf{p} \mathbf{b}^{pose}_t(\mathbf{p})$ denotes the most likely pose based on $\mathbf{b}^{pose}_{t}$, and $\varepsilon$ is a threshold. Note that instead of minimizing the pose error, we only limit the pose error below a threshold for navigation efficiency.

\section{Approach}
We begin this section by introducing the key ingredients of
our reinforcement-learning framework. Next, we describe the details of the architecture of our policy network, which works with a third-party localization module. Finally, we present the algorithm and the scenarios for training, along with the uncertainty settings in the simulation. 

\subsection{Reinforcement Learning Ingredients}
\subsubsection{\bf \emph{State space}}
A state $\mathbf{b}_{t}$ at time step $t$ consists of the probability distribution of the relative goal position~$\mathbf{b}_{t}^{goal}$, the readings of the
2D laser range finder $\mathbf{o}_{t}^{scan}$, and the three-channel pedestrian map $\mathbf{o}^{ped}$. 
In specific, $\mathbf{b}_{t}^{goal} = (x_t, y_t, a_t, \sigma_t^x, \sigma_t^y, \sigma_t^a)$ consists of the estimated relative target position $(x_t, y_t)$, the estimated relative target orientation $\alpha_t$, and  $(\sigma_t^x, \sigma_t^y, \sigma_t^\alpha)$ are respectively the estimated variance of $x_t$, $y_t$ and $\alpha_t$, which indicate the localization module's confidence (or uncertainty) in pose estimation~\cite{fox2003bayesian}.
$\mathbf{o}_t^{scan}$ are the measurements of the last frame from a
180-degree laser scanner that provides 720 distance values per
scanning (i.e., $\mathbf{o}_{t}^{scan} \in \mathbb{R}^{1 \times 720}$).
$\mathbf{o}_t^{ped}$ is a local grid map with three
channels (i.e., $\mathbf{o}_{t}^{ped} \in \mathbb{R}^{3 \times 48 \times 48}$). As specified in~\cite{yao2021crowd}, the first channel uses occupied grids to specify the locations and the shapes of surrounding pedestrians. The rest two channels use the value in the occupied grids to specify the speeds of corresponding pedestrians for the x-axis and y-axis, respectively. These motion data can be accessed directly in the simulator and estimated in the real world with human tracking.

\subsubsection{\bf \emph{Action space}}
The action space is a set of permissible
velocities in discrete space. The action of a differential robot
includes the translational and rotational velocity, i.e., $\mathbf{a}_t =
(v_t, \omega_t)$. In this work, considering the real robots' kinematics
and the real-world applications, we set the range of the
translational velocity $v \in \{0.0, 0.2, 0.4, 0.6\}$ and the rotational
velocity in $w \in \{-0.9, -0.6, -0.3, 0.0, 0.3, 0.6, 0.9\}$.

\subsubsection{\bf \emph{Reward function}}
The reward $\rho_{t}$ consists of six components, i.e.,
	{\small
		\begin{align*}
			\rho_{t} &=\rho^{app}_{t}+\rho^{pose}+\rho^{arr}_{t}+\rho^{col}_{t}+\rho_t^{lost}+\rho^{step}_{t},\\
			\rho^{app}_{t} &=k_{a}\left(\left\|\mathbf{l}_{t-1}^{gt}-\mathbf{l}_g\right\|-\left\|\mathbf{l}^{gt}_{t}-\mathbf{l}_g\right\|\right)/\left( 1 + \sigma_t^x + \sigma_t^y + \sigma_t^a \right), \\
			\rho^{pose}_t 
			&=k_{p} \left(\left\|\mathbf{p}_{t-1}-\mathbf{p}_{t-1}^{gt}\right\|-\left\|\mathbf{p}_{t}-\mathbf{p}_t^{gt}\right\|\right), \\
			\rho^{arr}_{t} &=\left\{\begin{array}{ll}
				r_{arr}  & \text { if }\left\|\mathbf{l}_{t}-\mathbf{l}_{g}\right\|<\varepsilon_{a},\\
				0 & \text{ otherwise,}
			\end{array}\right.\\
			\rho^{coll}_{t} &=\left\{\begin{array}{ll}
				r_{col} & \text { if collision}, \\
				0 & \text { otherwise,}
			\end{array}\right.\\
			\rho^{lost}_{t} &=\left\{\begin{array}{ll}
				r_{lost} & \text { if} \left\|\mathbf{l}_{t}-\mathbf{l}_{t}^{gt}\right\|>\varepsilon_{l} \text{, or} \left|{yaw}_{t}-yaw_{t}^{gt}\right|>\varepsilon_{yaw},\\
				0 & \text { otherwise,}
			\end{array}\right.\\
			\rho^{step}_{t} &= r_{step},
	\end{align*}}%
where $\mathbf{p}_{t}$ denotes the estimated pose of the robot at the time step $t$, 
$\mathbf{p}_{t}$ consists of the estimated location $\mathbf{l}_t$ and the estimated yaw angle $yaw_t$,
$\mathbf{p}_{t}^{gt}$ denotes the ground truth pose of the robot at the time step $t$, 
$\mathbf{p}_{t}^{gt}$ also consists of $\mathbf{l}_t^{gt}$ and $yaw_{t}^{gt}$, 
$\mathbf{l}_g$ denotes the location of the goal,
$k_{a}$, $k_{p}$, $r_{arr}$, $r_{col}$, $r_{lost}$, $r_{step}$, $\varepsilon_{a}$, $\varepsilon_{l}$, and $\varepsilon_{yaw}$ are hyperparameters.

The reward functions defined above provide both sparse and dense feedback for agents. The sparse feedback includes $\rho^{arr}_{t}$ for rewarding the agents arriving at the target, $\rho_{col}$ for penalizing collisions, and $\rho_{lost}$ for penalizing getting lost, these two types of feedback regulate the robots to satisfy the hard constraints. On the other hand, the dense feedback encourages the robots to approach the goal when the localization uncertainty is low (i.e., $\rho^{app}_{t}$) and to increase the localization accuracy (i.e., $\rho^{pose}_{t}$), and penalize the time consumption (i.e., $\rho^{step}$), this feedback adjusts the behavior of the robots to satisfy soft constraints.
As hard constraints are often more crucial than soft constraints, we assign large weights for hard constraints and small weights for soft constraints, i.e, $k_{a}=200$, $k_{p}=400$, $r_{arr}=500$, $r_{col} = -800$, $r_{lost}=-500$, $r_{step}<-5$, $\varepsilon_{a}=0.5m$, $\varepsilon_{l}=2.0m$, and $\varepsilon_{yaw}=0.25\pi$.

\begin{figure*}[t]
	\centering
	\subfigure[Hybrid]{
		\label{fig3:subfig:a} 
		\includegraphics[width=0.211\linewidth]{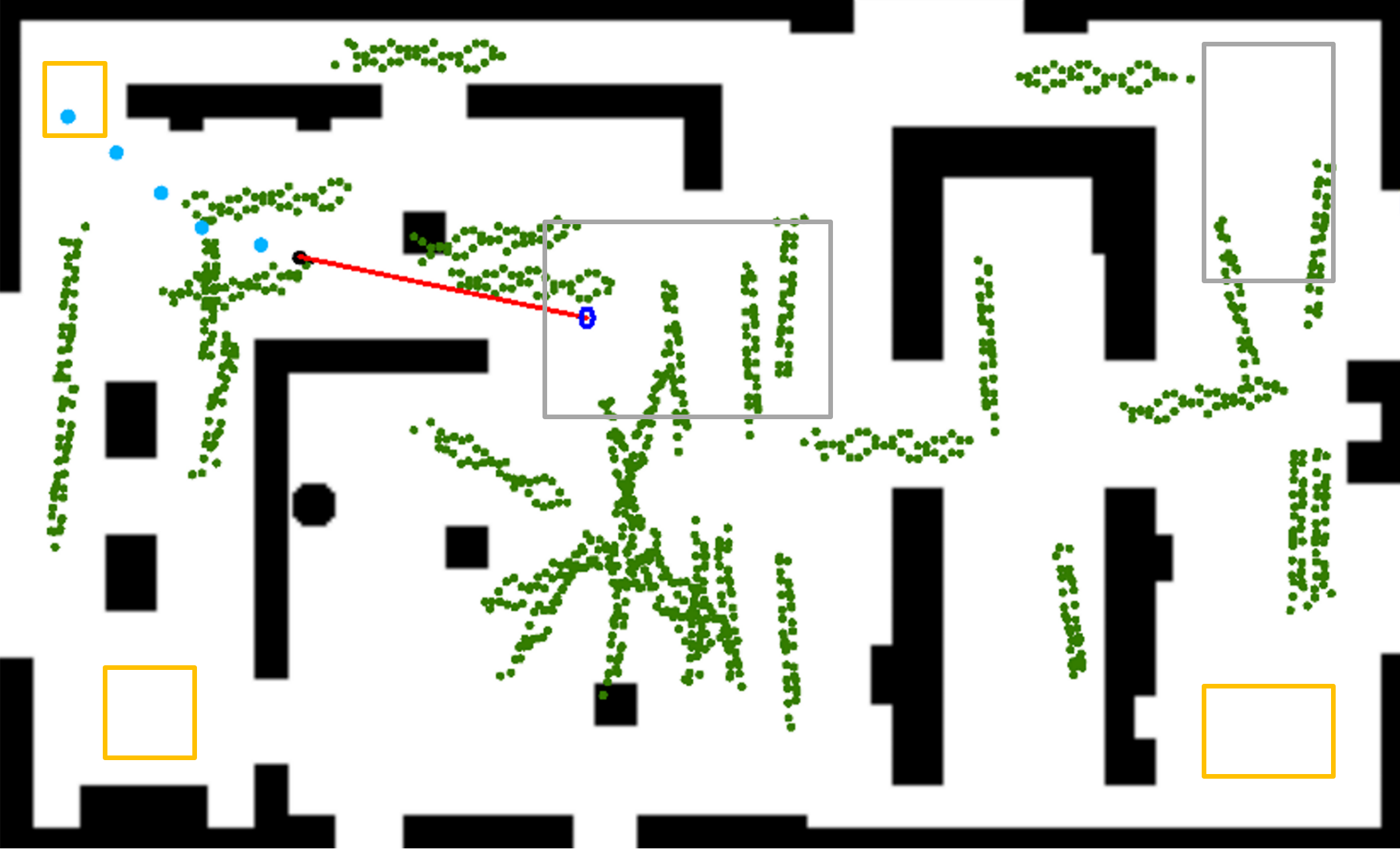}
	}
	\subfigure[Sparse]{
		\label{fig3:subfig:b}
		\includegraphics[width=0.211\linewidth]{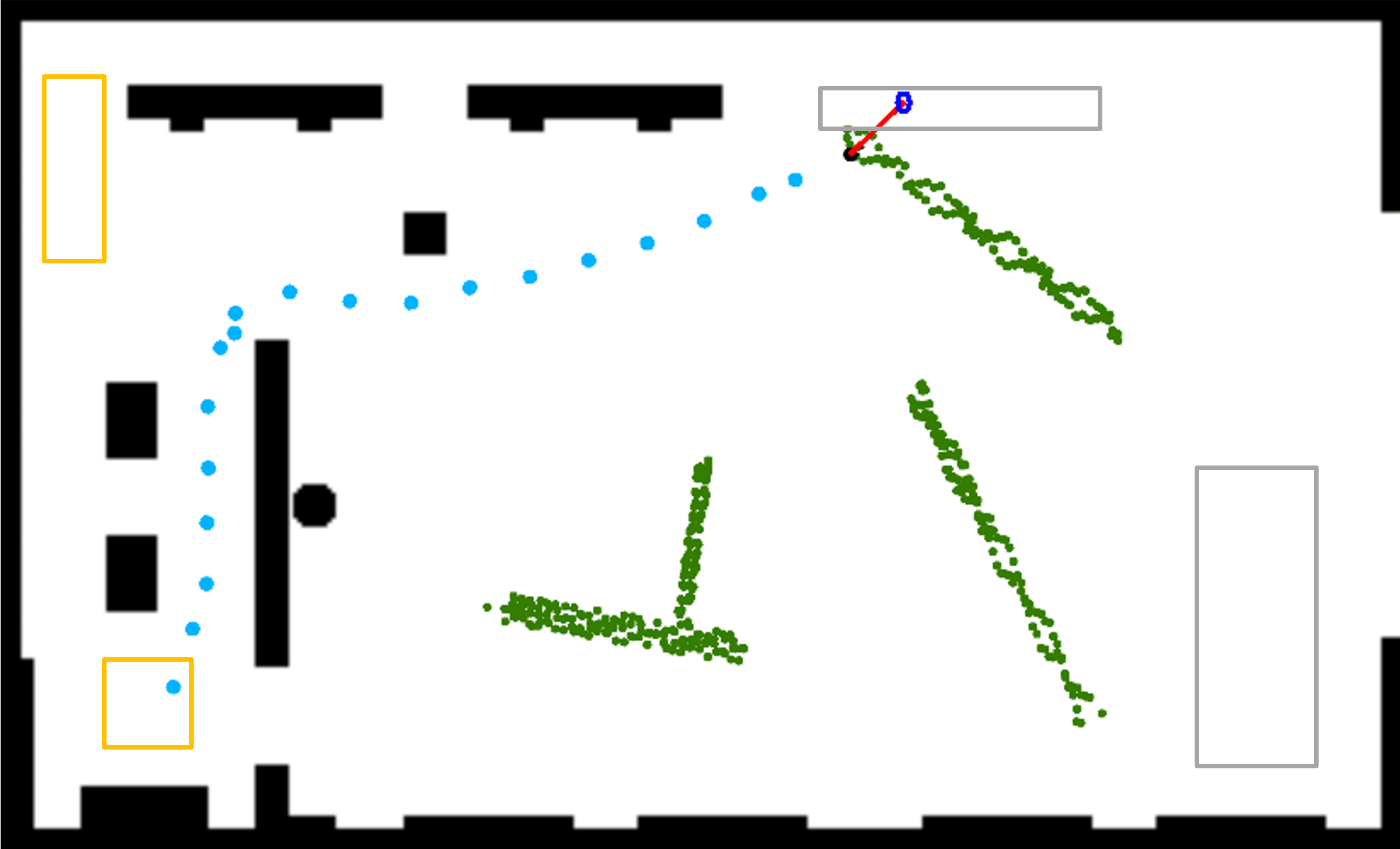}
	}
	\subfigure[Corridor]{
		\label{fig3:subfig:c} 
		\includegraphics[width=0.192\linewidth]{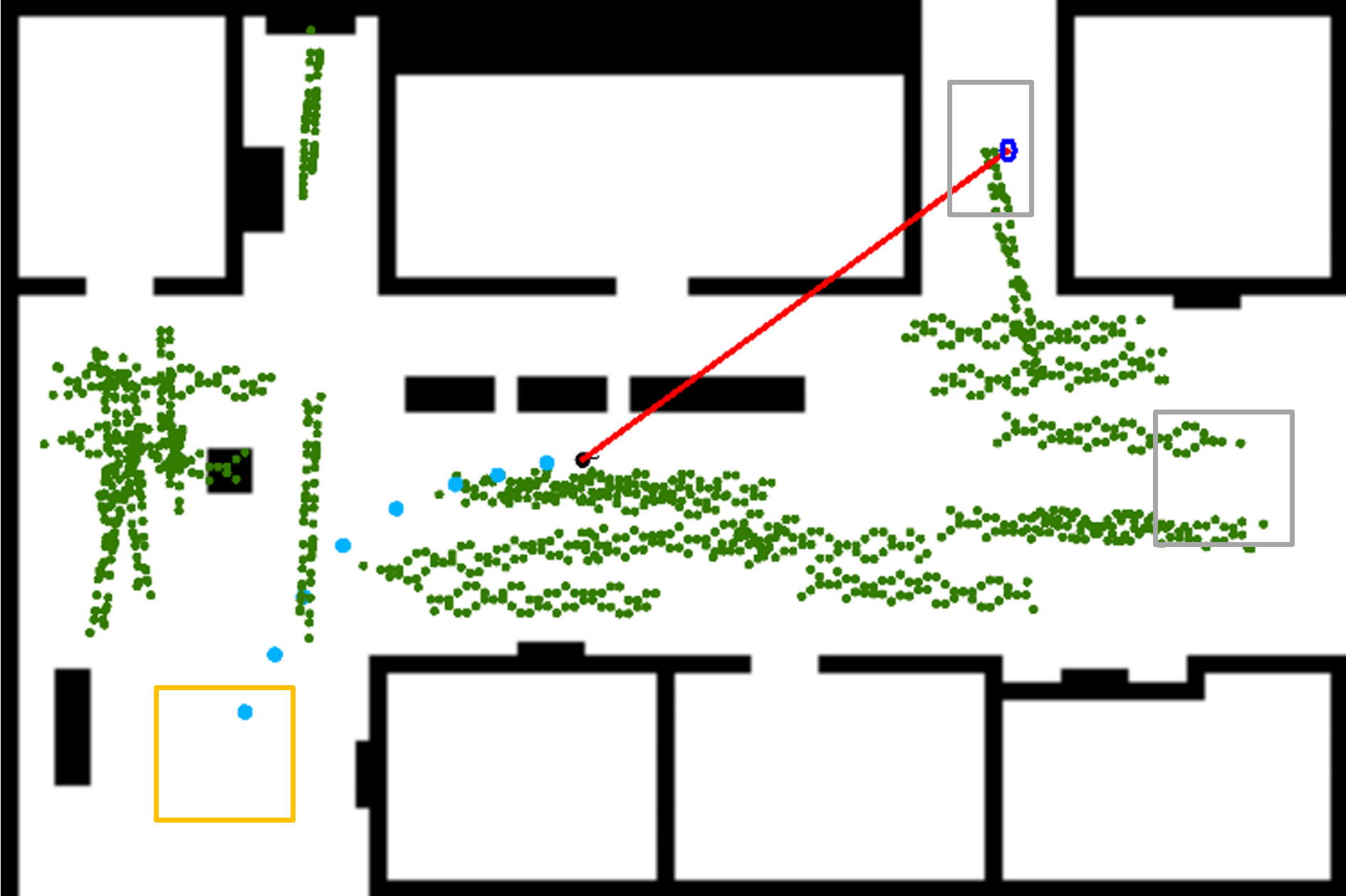}
	}
	\subfigure[Hall]{
		\label{fig3:subfig:d}
		\includegraphics[width=0.128\linewidth]{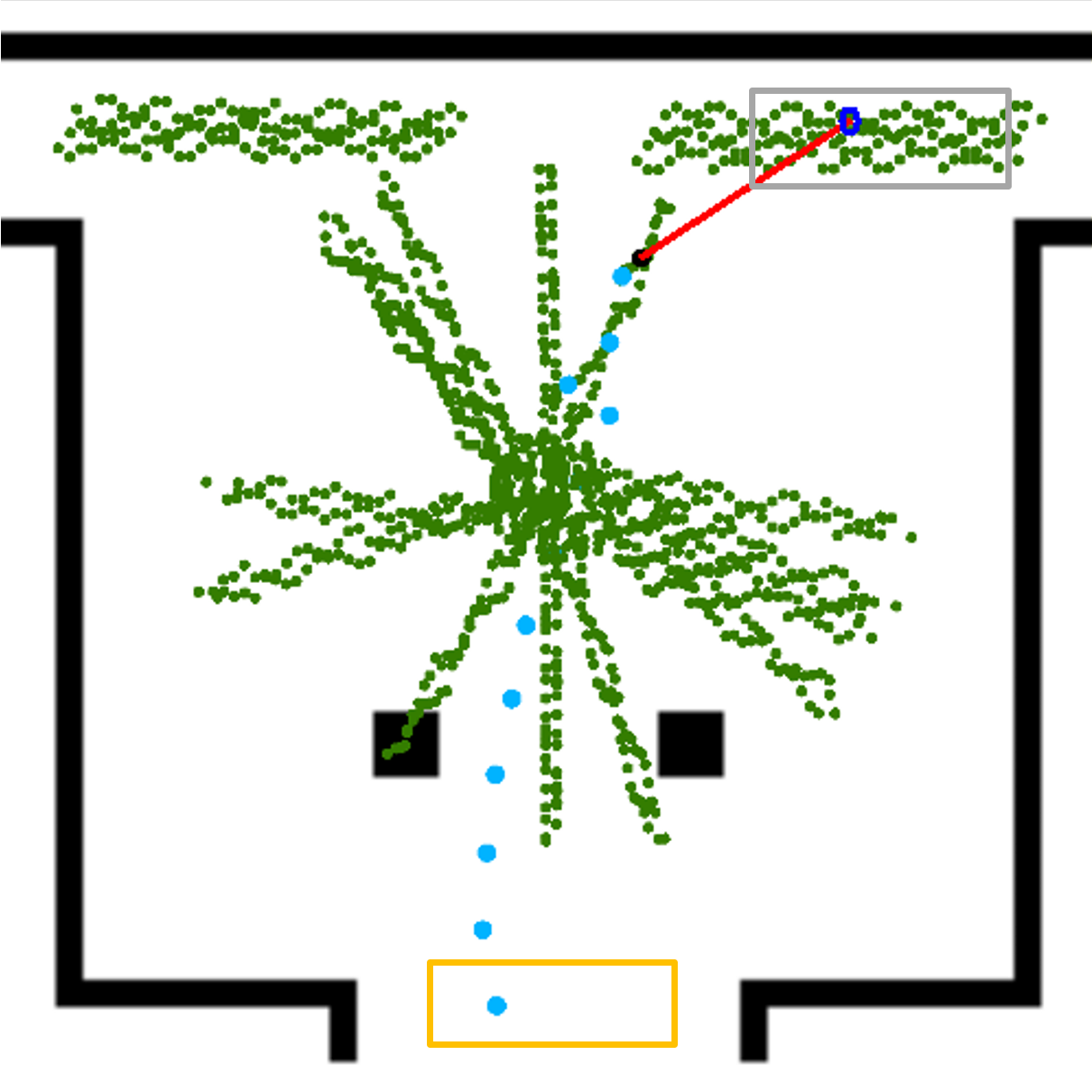}
	}
 	\subfigure[Irregular]{
		\label{fig3:subfig:e}
		\includegraphics[width=0.128\linewidth]{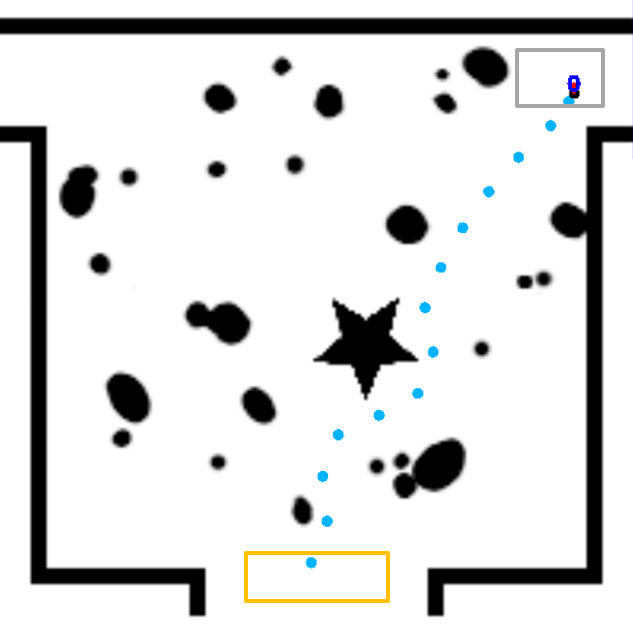}
	}
	\caption{We consider five scenarios in our experiments and their global top views are shown as above, where the red lines connect the robot to the target, blue dots denote the navigation paths of the robot, and green spots denote the legs' trajectories of pedestrians, the black blocks denote static obstacles, the yellow boxes denote the random region of the starting position, and the grey boxes denote the random region of the target position.}
	\label{fig3}
\end{figure*}

\subsection{Network Architecture}
As shown in Fig.~\ref{fig2}, we define a deep
convolutional neural network that embeds a pedestrian map encoder, a scan encoder, and a target encoder to compute the probability of choosing each action. 
In specific, the pedestrian map encoder employs three 
convolutional layers and a fully connected (FC) layer to extract the high-level features from the pedestrian maps.
The first layer convolves 64 two-dimensional filters with kernel size~=~$3\times3$ (Conv $3\times3$), stride~=~1
over the three channels of the pedestrian maps followed by ReLU nonlinearities (which are used by every layer in the entire network except the last layer).
The second and third layers have the same kernel size and stride as the first layer, but the second layer has 128 filters and the third layer has 256 filters.
The last FC layer has 512 rectifier units.
The scan encoder employs two convolutional layers and an FC layer to extract the high-level features from scan measurements. The first layer convolves 32 one-dimensional filters (Conv1D) with kernel size~=~5, stride~=~3 over the scans. The second layer convolves 32 one-dimensional filters with kernel size~=~3, stride~=~2. The FC layer has 512 rectifier units.
The target encoder has only one FC layer with 64 rectifier units, which transforms the estimated relative position and uncertainty of the target into a high-dimensional feature vector.
Then the network combines both the encoded vectors and feeds them to an FC layer with 512 units. At last, the network uses a softmax layer with 28
units to choose the pair of linear and angular velocities from
corresponding values.

\subsection{Training under Localization Uncertainty}
\subsubsection{\bf \emph{Training algorithm}}
To train the complex localizability-enhanced policy with multiple optimization items in the reward function, we need a policy gradient method that strikes a balance between ease of tuning, sample complexity, and sample efficiency. Therefore, we adopt proximal policy optimization (PPO)~\cite{schulman2017proximal}. 
The state-value function in PPO is approximated with a neural
network, which has the same architecture as our policy network, except that it has only one linear activation
unit in its last layer. 

\subsubsection{\bf \emph{Training Scenario}}~\label{ts}
We train the robot in our customized simulator~\cite{chen2020distributed}.
As shown in Fig.~\ref{fig3:subfig:a}, the \emph{hybrid} training scenario has a nonuniform density of spatial features and diverse pedestrian distribution. The scenario's size is $33\,m \times 20\,m$. The pedestrians in the simulator are driven by optimal reciprocal collision avoidance (ORCA)~\cite{ORCA}, or social force
model (SFM)~\cite{SFM} as~\cite{yao2021crowd}. The starting and target positions of pedestrians have small random ranges, which can generate various collision avoidance situations.

\subsubsection{\bf \emph{Uncertainty settings}}
Since we only need to obtain the relative position of the target point rather than the global position in our task, both map-based and mapless localization algorithms can be used.
The uncertainty of localization is mostly derived from the uncertainty of perception and travel distance measurement.
To better simulate the real world, we add perception uncertainty to the simulated laser range scanner and add the distance measurement uncertainty to the simulated odometer.

The perception uncertainty is simulated by \emph{Beam range finder model}~\cite{thrun2002probabilistic}, which can be parameterized with $\{z_{hit}, z_{short}, z_{max}, z_{rand}, \sigma_{hit}\}$.
\emph{Beam range finder model}	
incorporates four types of measurement errors: small measurement noise (with weight $z_{hit}$ and Gaussian noisy $\mathcal{N}\left(0, \sigma_{\mathrm{hit}}^2\right)$), errors due to unexpected objects (with weight $z_{short}$),
errors due to failures to detect objects (with weight $z_{max}$), and random unexplained noise (with weight $z_{rand}$). 
As we have pedestrians in the simulation, we can directly generate errors from unexpected objects. Therefore, we set $z_{short}=0$, $z_{hit}=0.98$, $z_{max}=0.01$, $z_{rand}=0.01$, and $\sigma_{\mathrm{hit}}=0.02$.
We assume that the odometer is calibrated well and has no systematic error. Hence, we only add non-systematic errors to the simulated odometer with dynamics randomization as~\cite{peng2018sim,choi2019deep} by multiplying the $a_t$ with $\mathcal{N}\left(1, 0.01\right)$.

\begin{table*}[bhp]
	\centering
	\caption{Performance metrics (as mean/standard deviation) evaluated for different methods.}
            \begin{tabular}{lrrrrrrrrr}
            \toprule
            Method & \multicolumn{1}{l}{\#metrics} &       &       &       &       &       &       &       &  \\
        \cmidrule{2-10}          & \multicolumn{1}{l}{AR} & \multicolumn{1}{l}{CR} & \multicolumn{1}{l}{LR} & \multicolumn{1}{l}{SR} & \multicolumn{1}{l}{$\bar{t}$ (mean/std)} & \multicolumn{1}{l}{$\bar{d}$ (mean/std)} & \multicolumn{1}{l}{$\overline{e_p}$ (mean/std)} & \multicolumn{1}{l}{$\overline{e_\alpha}$ (mean/std)} & \multicolumn{1}{l}{$\overline{\sum_{v}}$ (mean/std)} \\
            \midrule
            DWA   & 0.631 & 0.322 & 0.035 & 0.012 & 23.82/7.806 & \textbf{11.19/1.80} & 0.3083/0.3524 & \textbf{0.1257/0.1678} & 0.5144/2.8543 \\
            DRL\_laser & 0.542 & 0.395 & 0.057 & 0.006 & 29.88/10.76 & 13.09/2.83 & 0.3265/0.6434 & 0.1928/0.2169 & 0.3018/2.6280 \\
            DRL\_laser+ped & 0.641 & 0.153 & 0.054 & 0.152 & 30.76/12.79 & 16.18/5.57 & 0.3261/0.6654 & 0.2697/0.2277 & 0.2951/2.9579 \\
            LNDRL(ours) & \textbf{0.892} & \textbf{0.104} & \textbf{0.003} & \textbf{0.001} & \textbf{22.86/5.26} & 12.21/2.15 & \textbf{0.3071/0.2385} & 0.1887/0.1775 & \textbf{0.1062/1.1299} \\
            \midrule
            -variance & 0.790 & 0.188 & 0.021 & 0.001     & 23.03/3.87 & 12.49/1.79 & 0.3072/0.3821 & 0.2338/0.2189 & 0.2301/1.8215 \\
            -reward & 0.743 & 0.227 & 0.030 & 0.000 & 23.54/4.97 & 12.69/2.25 & 0.3725/0.5212 & 0.1873/0.2111 & 0.3153/2.7664 \\
            \bottomrule
            \end{tabular}
	\label{table2}
\end{table*}

\section{Experiments}
In this section, we will begin by briefly discussing our implementation details. Next, we will define the evaluation metrics and introduce the testing scenarios. Afterward, we will evaluate the performance of our method in various simulation scenarios, comparing it with other approaches. To specifically illustrate the consideration of localizability, we will provide examples of trajectories generated by our method. In addition, we will display the activation of the first laser feature layer to demonstrate the extracted geometry features that contribute to enhanced localizability decisions. Ablation studies will then be conducted to show the role of our augmented state and reward metric. Finally, we will provide details regarding our deployment of the trained policy on a physical robot.

\subsection{Implementation Details}
We use adaptive Monte Carlo localization (AMCL)~\footnote{http://wiki.ros.org/amcl} for localization, as it can efficiently quantify the uncertainty of localization results with a covariance matrix, which is very important for training. AMCL receives $10\,Hz$ laser data with a maximum range of $12m$ and $10\,Hz$ odometer data in the simulation, consistent with the sensors we use in the real world.
The detailed AMCL parameters can be found in the video link.
We train the localizability-enhanced policy following the PPO algorithm with the same hyperparameters as~\cite{yao2021crowd}. 
Both the policy and value networks are implemented in Pytorch~\cite{paszke2017automatic} and trained with the Adam optimizer~\cite{kingma2014adam} on a computer with AMD R9 5950X CPU and Nvidia RTX 3090 GPU. 
It takes around 16 hours to run about $1.2\times10^6$ steps in PPO when the networks converge well in the training scenario.

\subsection{Metrics and Testing Scenarios}
We use the following metrics to evaluate the performance:
\begin{itemize}
	\item {\bf Arrival rate (AR)}: the ratio of the episodes that end with the robot reaching its target without any collisions.
	\item {\bf Collision rate (CR)}: the ratio of the episodes that collisions occur.
	\item {\bf Lost rate (LR)}: the ratio of the episodes that the robot gets lost (the threshold is defined in the reward function).
	\item {\bf Stuck rate (SR)}: the ratio of the episodes exceeding the maximum episode length.
	\item {\bf Reach time ($\bar{t}$)}: the average time required for the robot to successfully reach its target.
	\item {\bf Moving distance ($\bar{d}$)}: the average moving distance required for the robot to successfully reach their
	targets.
	\item {\bf Positioning error ($\overline{e_p}$)}: the average positioning error after each action is performed.
	\item {\bf Angular error ($\overline{e_\alpha}$)}: the average yaw error after each action is performed.
	\item {\bf Sum of pose variance ($\overline{\sum_{v}}$)}: the average sum of $\sigma^x_t$, $\sigma^y_t$ and $\sigma^\alpha_t$ after each action is performed. 
\end{itemize}

Figure~\ref{fig3} illustrates five different scenarios that are considered in our experiments. Please note that we only use \emph{hybrid} scenario as the training environment and evaluate the performance in these five scenarios. 
The pedestrians in the testing scenarios are driven by the same algorithm as the training scenario, and also have small random regions for choosing start and target locations. As \emph{hybrid} scenario is introduced in section~\ref{ts}, we only introduce the other four scenarios here:
\begin{itemize}
	\item {\bf Sparse} scenario is $33\,m \times 20\,m$ with 4 pedestrians in it. It seems extremely simple, however, the robot may get lost in the middle open space due to the lack of spatial features.
	\item {\bf Corridor} scenario is $30\,m \times 20\,m$ with 22 pedestrians in it. The robot needs to interact with dense crowds of people heading in the same or the opposite direction.
	\item {\bf Hall} scenario is $20\,m \times 20\,m$ with 18 pedestrians in it. The robot needs to interact with pedestrians in all directions.
    \item {\bf Irregular} scenario is $20\,m \times 20\,m$ with irregular obstacles in it. The robot needs to adapt to unseen shapes of obstacles.
\end{itemize}

\subsection{Simulation Experiments}
\subsubsection{\bf \emph{Baselines}}
We compare our LNDRL (Lacalizability-enhanced Navigation with Deep Reinforcement Learning) with a traditional collision avoidance method and two DRL-based methods, all of which don't consider  localizability. For the sake of fair comparison, all the methods receive the same laser data and use the same localization algorithm with the same parameters to estimate the robot pose. To handle the unstructured raw laser data, we set Dynamic Window Approach (DWA)~\cite{fox1997dynamic} as the classical benchmark.
DRL\_laser is a DRL-based method with only raw laser data as the observation, its network architecture refers to \cite{fan2020distributed}, which directly predicts
the motions of nearby pedestrians from three consecutive frames of laser data. DRL\_laser+ped is an enhanced version of DRL\_laser, which adds pedestrian maps to the inputs of the network as LNDRL. The main difference between the two DRL-based methods and our LNDRL is that they assume localization is perfectly accurate. In specific, there is no $\sigma_t^x$, $\sigma_t^y$, and $\sigma_t^a$ in their state space, and no $\rho_t^{pose}$, $\rho_t^{lost}$ in their reward function. The other settings for the two DRL-based methods are the same as LNDRL. 

\subsubsection{\bf \emph{Evaluation results}}
For each method, we performed 500 tests under every scenario and calculate the mean of each metric. The test results are summarized in the top half of Table~\ref{table2}. LNDRL outperforms other methods in terms of lost rate and has the highest arrival rate. However, behaviors that prioritize localization can reduce navigation efficiency, resulting in a longer moving distance. Thus, LNDRL's performance in terms of this metric is not the best. Nevertheless, LNDRL strikes a balance between localization robustness and navigation efficiency, as it achieves the shortest reach time. We found that sometimes sacrificing part of the navigation efficiency for localization robustness can help the robot reach the target faster. Inaccurate localization can cause the robot to fail to reach the target and increase the stuck rate. A possible strategy is to temporarily trade off localization accuracy for efficiency as long as the pose error remains within acceptable thresholds. The penalty for increased pose error can be compensated later by reducing the pose error. This is why LNDRL does not significantly outperform the other methods in terms of average positioning and angular errors but has relatively low standard deviations on these two metrics. Moreover, LNDRL achieves the best performance in the sum of pose variance metric, indicating that it maintains a more confident localization algorithm most of the time.

In comparison with other methods, DWA achieves the shortest moving distance and smallest angular error. We attribute this to two factors: (1) DWA has little randomness and can head straight to the target in the exact direction. In contrast, due to the randomness in training and execution, DRL policies in our experiments may have slight oscillations in the travel direction. (2) DWA has weak intentions to avoid pedestrians, resulting in fewer turns and lower angular error. However, this strategy comes at the cost of a higher collision rate, which is similar to that of DRL\_laser. Although DRL\_laser has a smaller angular error, it has a higher collision rate than DRL\_laser+ped, indicating the benefits of incorporating pedestrian maps to avoid pedestrians.

\subsubsection{\bf \emph{Example trajectories}}
\begin{figure*}[tp]
\centering
\subfigure[DRL\_laser+ped in sparse]{
    \label{fig4:subfig:a} 
    \includegraphics[width=0.225\linewidth]{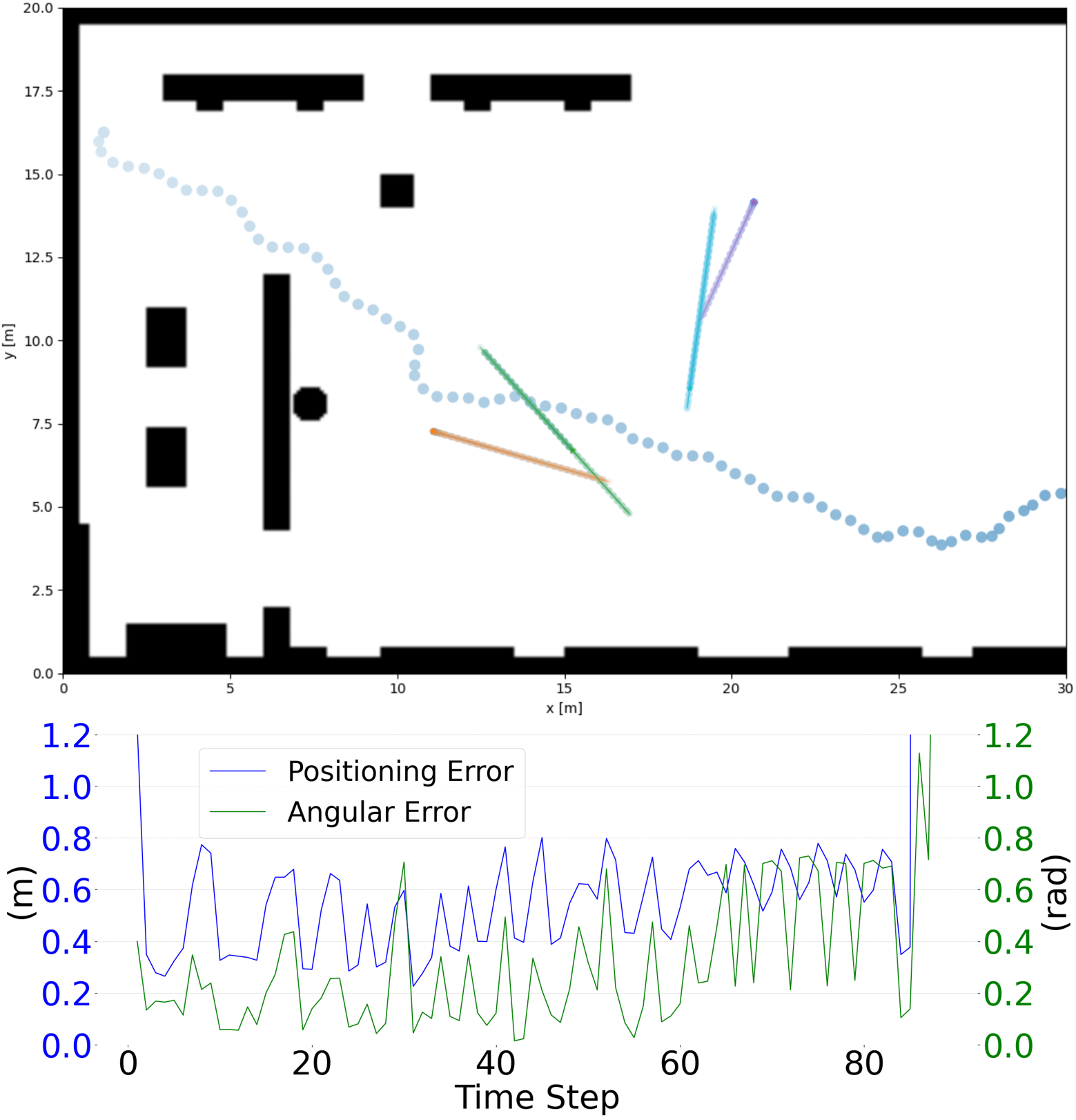}
}
\subfigure[LNDRL in sparse]{
    \label{fig4:subfig:b}
    \includegraphics[width=0.225\linewidth]{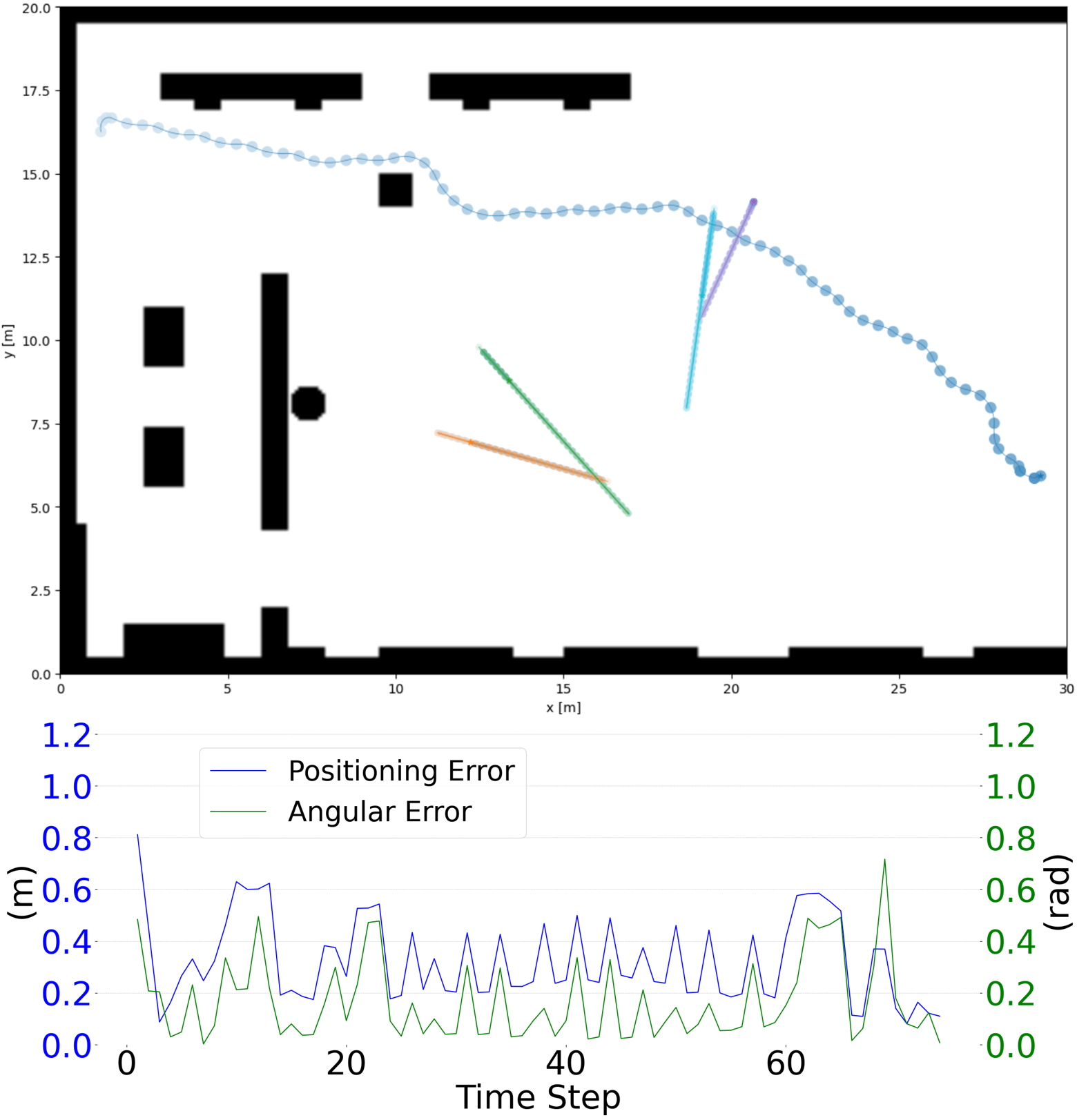}
}
\subfigure[DRL\_laser+ped in corridor]{
    \label{fig4:subfig:c} 
    \includegraphics[width=0.225\linewidth]{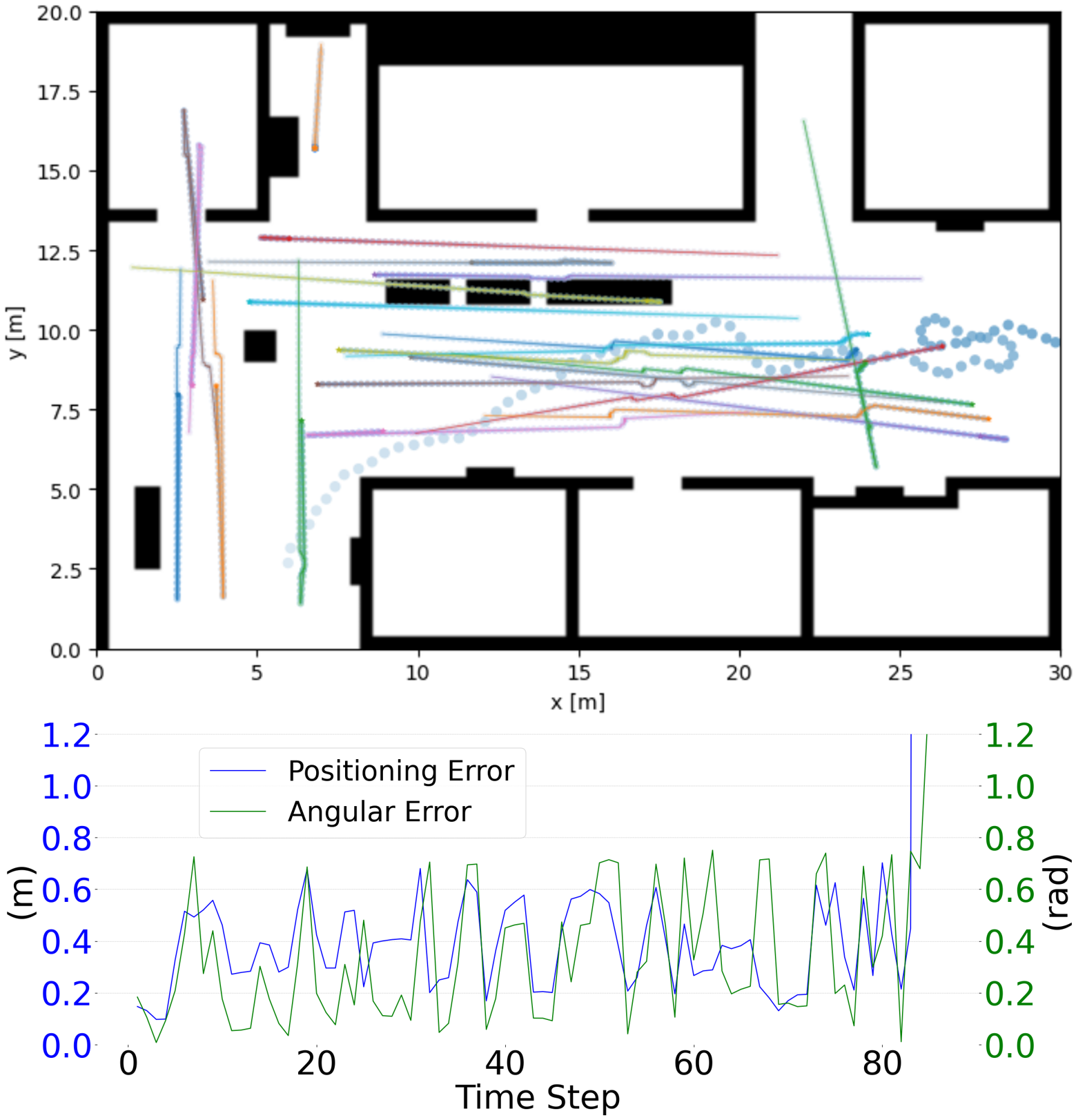}
}
\subfigure[LNDRL in corridor]{
    \label{fig4:subfig:d}
    \includegraphics[width=0.225\linewidth]{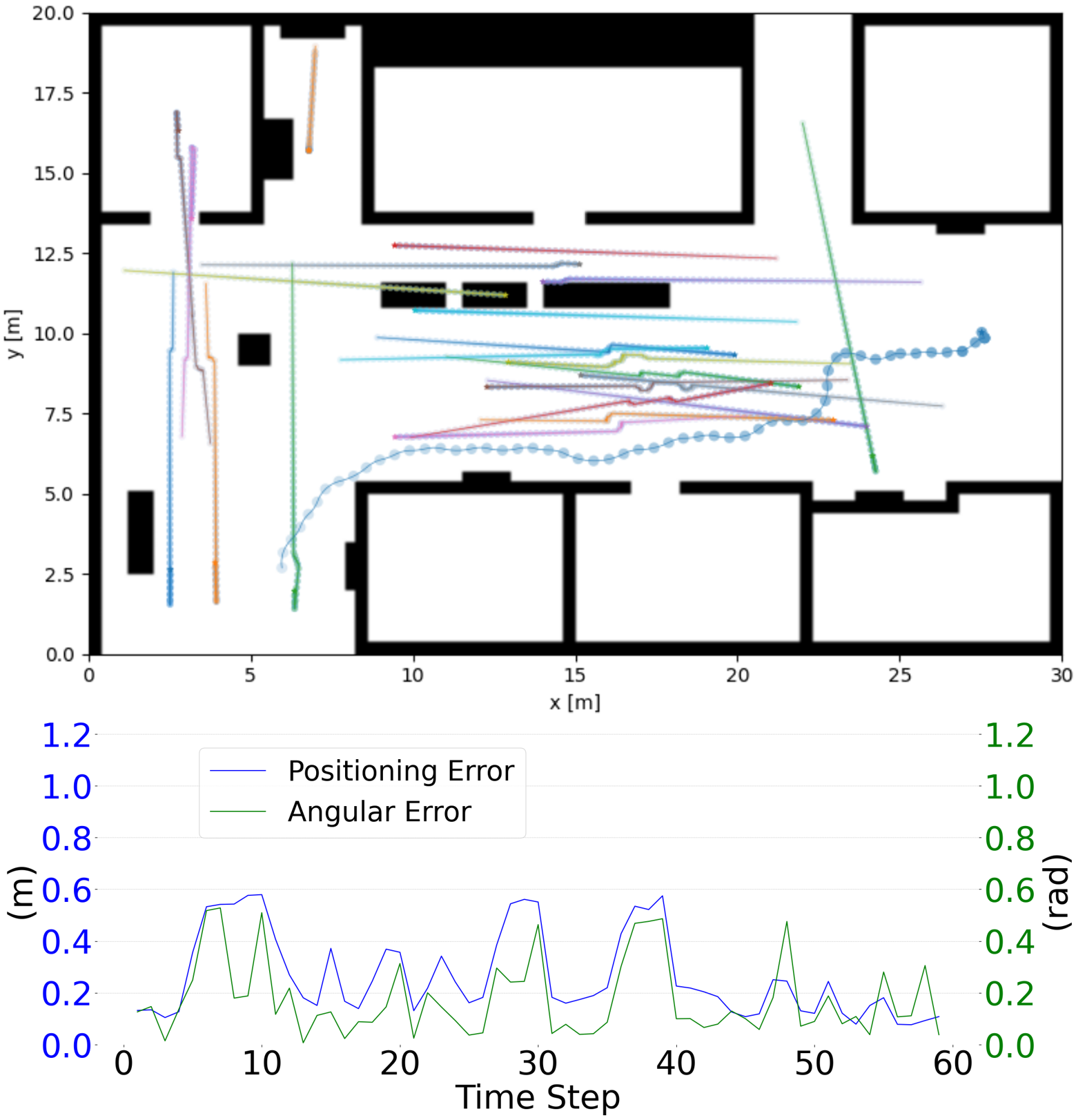}
}
\caption{Robot's trajectories generated by DRL\_laser+ped and LNDRL (ours) in the \emph{sparse} and \emph{corridor} scenarios. The blue spots denote the trajectories of the robot and other spots denote the trajectories of pedestrians.
    The line charts below show the positioning and angular error at each time step on the corresponding trajectories.
    We can observe that LNDRL prefers the routes with more spatial features to reach the target and constrains its pose error at the lower level.}
\label{fig4}
\end{figure*}

As shown in Fig.~\ref{fig4}, we illustrate the robot's trajectories generated by LNDRL and DRL\_laser+ped (which is the method that achieves the highest arrival rate among the other three methods). We can observe in the \emph{sparse} scenario that our LNDRL drives the robot moves along a path towards the goal with more geometry features than DRL\_laser+ped and limits the positioning and angular error in a relatively low range. In the \emph{corridor} scenario, our LNDRL takes a route close to the wall to reach the target, which helps the robot avoid being surrounded by pedestrians DRL\_laser+ped and get more accurate localization. An interesting detail is that after the robot finishes its route along the wall, instead of heading straight for the end, it first takes a look in the direction where the geometry features are denser, which significantly reduces its localization error.

\subsubsection{\bf \emph{Activation visualization}}

\begin{figure}[htp]
	\centering
	\subfigure[Scenario]{
		\label{fig6:subfig:a} 
		\includegraphics[width=0.36\linewidth]{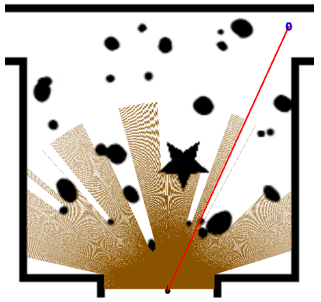}
	}
	\subfigure[Raw laser data]{
		\label{fig6:subfig:b}
		\includegraphics[width=0.57\linewidth]{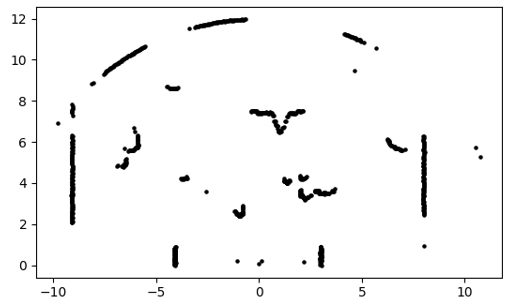}
	}
	\subfigure[Activation of the first laser feature layer]{
		\label{fig6:subfig:c} 
		\includegraphics[width=1.0\linewidth]{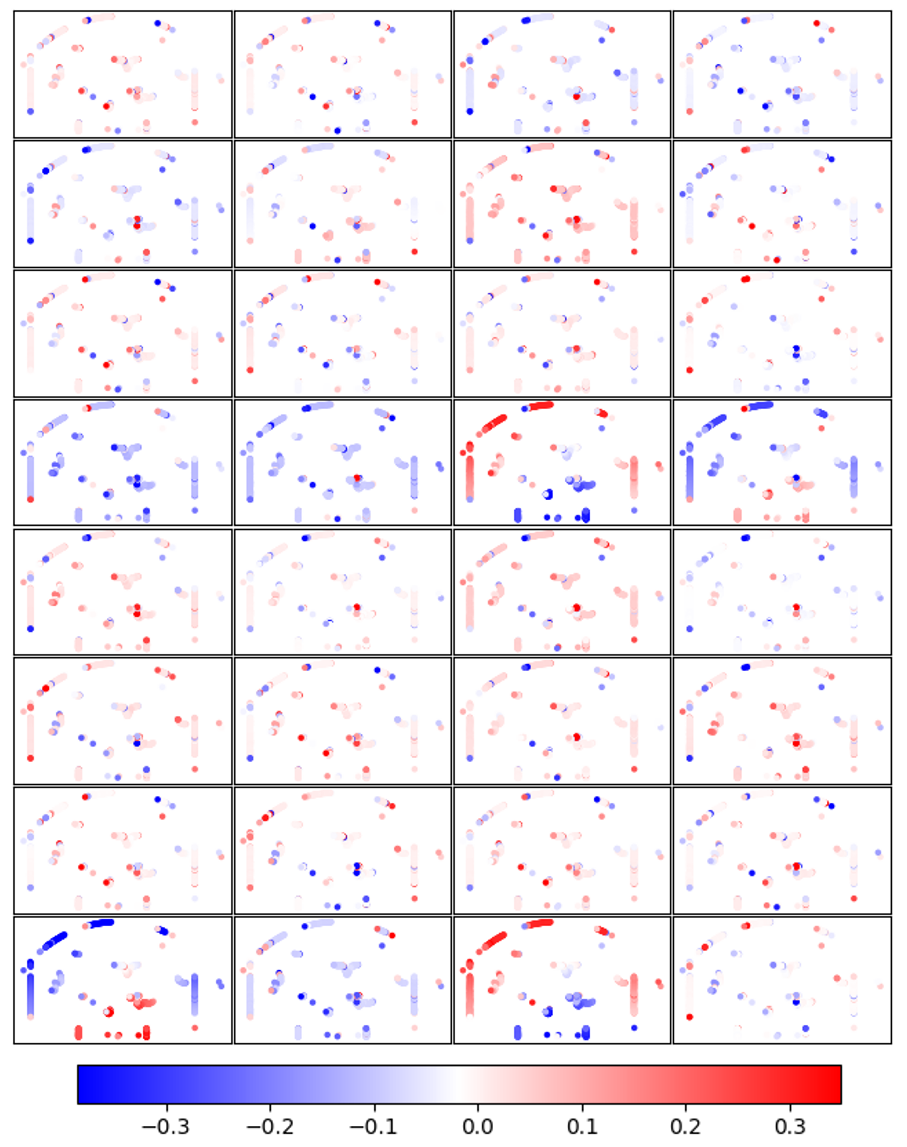}
	}
	\caption{ How the first layer of scan encoder “sees” the irregular obstacles. (a) an example scenario and the (b) raw laser data perceived by the robot in it. (c) Activation of the first laser feature layer with 32 1D filters.}
	\label{fig6}
\end{figure}

We create a visual representation of the numerical data output by the first 1D convolutional layer by assigning colors to different values and mapping each channel to its corresponding position in the original laser data. In Fig.~\ref{fig6:subfig:c}, the activation maps clearly show the corners of obstacles, which we believe to be important features for localization. Additionally, we observed features related to the distance from the robot, which may be useful for obstacle avoidance, since close obstacles are more urgent to avoid.

\subsubsection{\bf \emph{Ablation studies}}
We conduct ablation studies to show the role of our augmented state and reward metric.
In specific, two experiments were conducted: 
one with the removal of $\sigma_t^x$, $\sigma_t^y$, and $\sigma_t^a$ from the state space, denoted as `-variance', 
and another with the removal of $\rho_t^{pose}$ (which is the most distinct item in the reward function compared to other DRL navigation methods) from the reward function, denoted as `-reward'. 
The results of the ablation study can be found in the bottom half of Table~\ref{table2}, which reveals that both the incorporation of localization quality and the frequent feedback on localization are important. It is worth noting that while both components are related to localization, their removal significantly increases the collision rate.
Our guess for this is that, when less confident about localization, the robot sometimes regards the target as unreachable and thus tends to end the episode by collision so as to reduce the penalty in time consumption.

\subsection{Real World Experiments}
\begin{figure}[tp]
	\centerline{\includegraphics[width=1\linewidth]{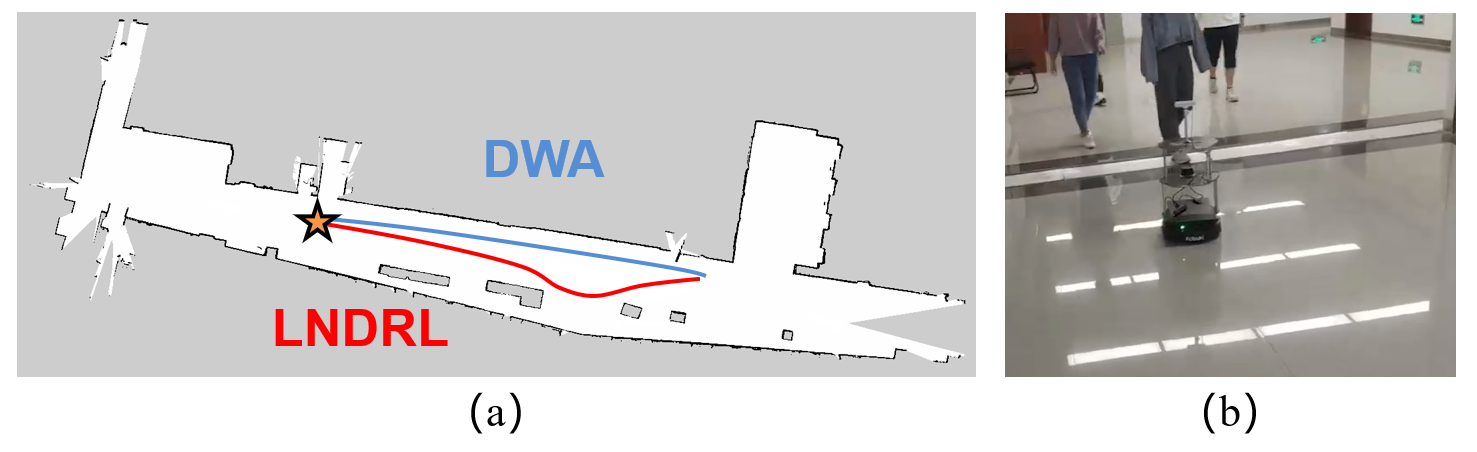}}
	\caption{(a) We compare LNDRL with DWA in a long corridor scenario ($80m \times 22m$), the five-pointed star represents the target point. (b) We use LNDRL to avoid pedestrians.}
	\label{fig5}
\end{figure}
To deploy the trained LNDRL model in a real-world environment, we equip a TurtleBot2 with an RPLIDAR-A2M12 laser range scanner and use a Realsense D455 depth camera as the pedestrian tracking sensor. We utilize the SPENCER people tracking framework~\cite{linder2016multi} to track the surrounding pedestrians and generate a pedestrian map. To estimate the robot's pose, we apply the Adaptive Monte Carlo Localization (AMCL) algorithm. We then evaluate LNDRL's performance in a long corridor scenario where glass walls are on one side and sofas on the other. As illustrated in Fig.~\ref{fig5}, our approach enables the robot to navigate close to the sofas and thus more fully use geometric features for more accurate localization.
Furthermore, LNDRL can also avoid pedestrians in the real world.
The video can be found at  \url{https://github.com/YohannnChen/LNDRL}.

\section{Conclusions}

In this paper, we present a novel approach to localizability-enhanced navigation in dynamic human environments using deep reinforcement learning. Our proposed method addresses the limitations of prior navigation methods by automatically extracting and assigning importance to geometric features that help with laser localization. It balances navigation efficiency, obstacle avoidance safety, and localization robustness by leveraging trial and error. We have also introduced two techniques, an augmented state representation and a reward metric, that facilitate the planner's learning. Our experimental results demonstrate significant improvements in lost rate and arrival rate across various simulation environments. Furthermore, our method generalizes well for real-world deployment on a physical robot. 

Our limitations are mainly in the following two aspects.
Firstly, we have not considered the acceleration limits of
physical robots, which could lead to a gap between simulation and real-world experiments. Secondly, more stringent collision-free guarantees are needed to be provided to further improve safety for navigation in human environments. Therefore, we will try to solve
these two problems in our future work.



\bibliographystyle{IEEEtran}
\bibliography{IEEEabrv,root}

\end{document}